\pdfoutput=1
\documentclass[11pt]{article}

\PassOptionsToPackage{table}{xcolor}
\usepackage[preprint]{acl}

\usepackage{times}
\usepackage{latexsym}

\usepackage[T1]{fontenc}
\usepackage[utf8]{inputenc}

\usepackage{microtype}

\usepackage{inconsolata}

\usepackage{graphicx}

\usepackage{url}
\usepackage{booktabs}
\usepackage{nicefrac}
\usepackage{amsmath,amsfonts,amssymb,amsthm}

\usepackage[export]{adjustbox}
\usepackage{wrapfig}
\usepackage{soul}
\usepackage{makecell}
\usepackage{multirow}
\usepackage{balance}

\usepackage{xspace}
\usepackage[english]{babel}
\usepackage{bm}
\usepackage{xurl}
\usepackage{algorithm,algorithmic}

\usepackage{arydshln}

\usepackage{tcolorbox}
\usepackage{soul}

\newcommand{\method}{RLAE\xspace}

\newcommand{\methodppo}{RLAE$_\text{PPO}$\xspace}
\newcommand{\methodmappo}{RLAE$_\text{MAPPO}$\xspace}

\title{\raisebox{-1ex}{\includegraphics[width=0.05\textwidth]{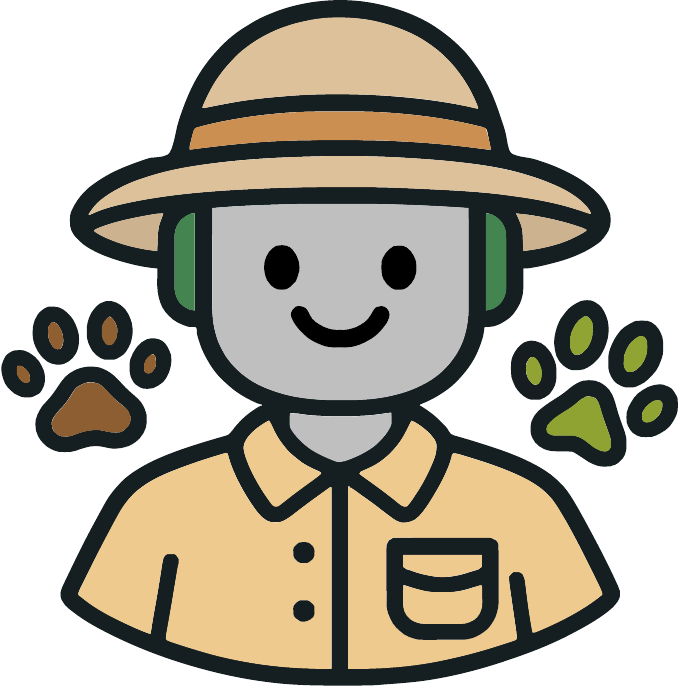}} \method: Reinforcement Learning-Assisted Ensemble for LLMs}

\author{
 \textbf{Yuqian Fu\textsuperscript{1,2}},
 \textbf{Yuanheng Zhu\textsuperscript{1,2}},
 \textbf{Jiajun Chai\textsuperscript{1,2,3}},\\
 \textbf{Guojun Yin\textsuperscript{3}},
 \textbf{Wei Lin\textsuperscript{3}},
 \textbf{Qichao Zhang\textsuperscript{1,2}},
 \textbf{Dongbin Zhao\textsuperscript{1,2}}
\\
\\
 \textsuperscript{1}Institute of Automation, Chinese Academy of Sciences,\\
 \textsuperscript{2}School of Artificial Intelligence, University of Chinese Academy of Sciences,\\
 \textsuperscript{3}Meituan
\\
 \small{
   {\texttt{\{fuyuqian2022,yuanheng.zhu\}@ia.ac.cn}}
 }
}

\begin{document}
\maketitle

\begin{abstract}
Ensembling large language models (LLMs) can effectively combine diverse strengths of different models, offering a promising approach to enhance performance across various tasks.
However, existing methods typically rely on fixed weighting strategies that fail to adapt to the dynamic, context-dependent characteristics of LLM capabilities.
In this work, we propose \textit{\textbf{R}einforcement \textbf{L}earning-\textbf{A}ssisted \textbf{E}nsemble for LLMs} (\method), a novel framework that reformulates LLM ensemble through the lens of a Markov Decision Process (MDP).
Our approach introduces a RL agent that dynamically adjusts ensemble weights by considering both input context and intermediate generation states, with the agent being trained using rewards that directly correspond to the quality of final outputs.
We implement \method using both single-agent and multi-agent reinforcement learning algorithms (\methodppo and \methodmappo), demonstrating substantial improvements over conventional ensemble methods.
Extensive evaluations on a diverse set of tasks show that \method outperforms existing approaches by up to $3.3\%$ accuracy points, offering a more effective framework for LLM ensembling.
Furthermore, our method exhibits superior generalization capabilities across different tasks without the need for retraining, while simultaneously achieving lower time latency.
\end{abstract}

\section{Introduction}

Recent years have witnessed remarkable progress in large language models (LLMs), demonstrating impressive capabilities in natural language understanding and reasoning. 
However, significant performance variations persist among different models in downstream tasks due to three primary factors: training data bias, architectural differences, and diversity in training algorithms~\citep{jiang-etal-2023-llm}. 
For example, GPT-4o~\citep{achiam2023gpt} excels in mathematical reasoning, while Claude-series models~\citep{claude2024} demonstrate superior performance in code generation tasks. 
Nevertheless, retraining these models with mixed training data, integrating model architectures, or adjusting training algorithms are expensive and impractical.
Therefore, these distinct capabilities motivate research into LLM ensemble methods that leverage the diverse strengths of different models~\citep{chen2025harnessing,frick2025prompt,lu2024merge}, aiming to achieve enhanced performance on given tasks.

\begin{figure}
    \centering
    \includegraphics[width=\linewidth]{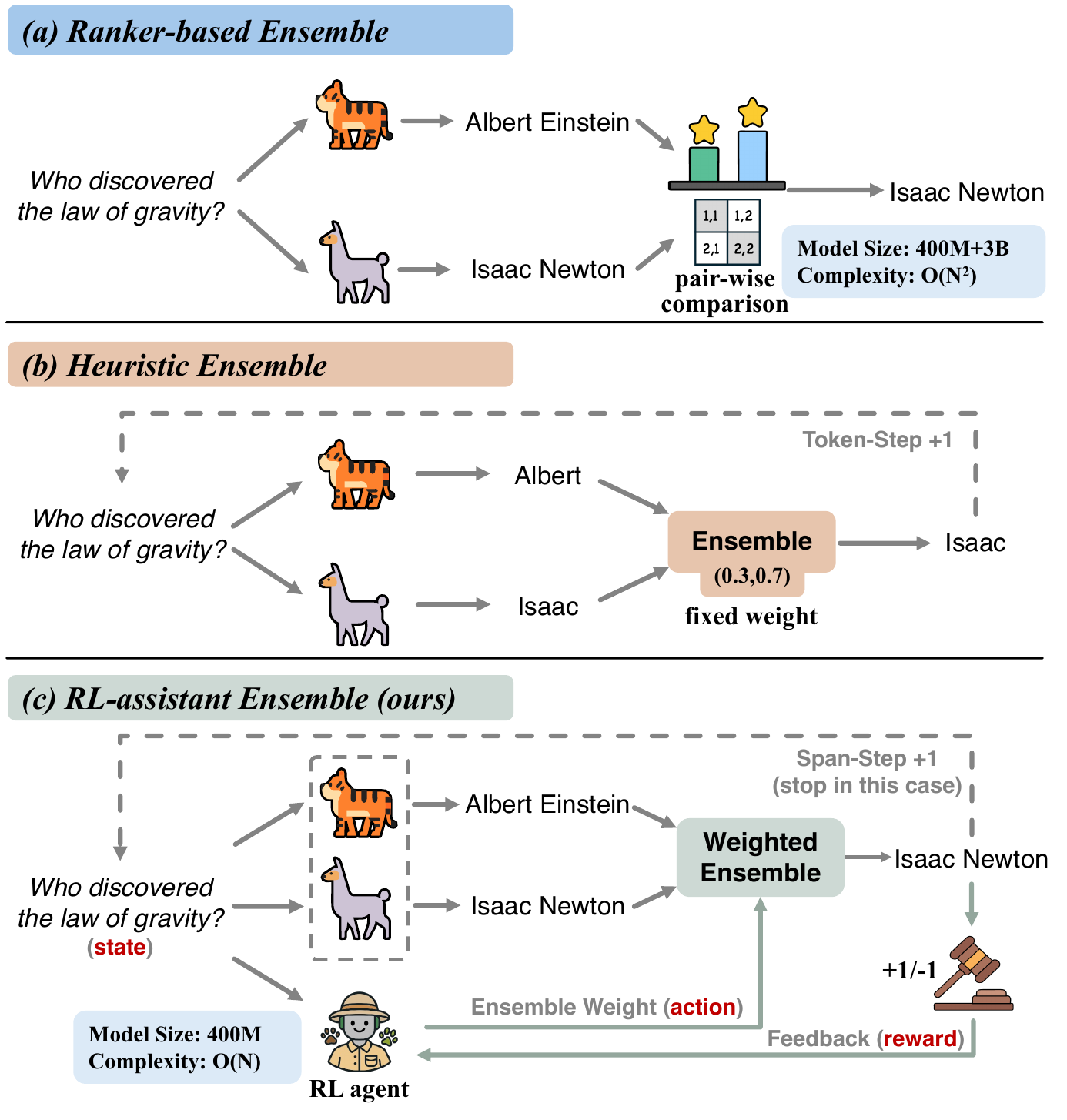}
    \caption{Overview of Ranker-based Ensemble, Heuristic Ensemble, and our RL-assisted Ensemble methods.}
    \label{fig:motivation}
    \vspace{-1.5em}
\end{figure}

The complexity of LLMs presents unique challenges for ensemble, requiring a more flexible framework beyond previous approaches.
Figure~\ref{fig:motivation} illustrates the key differences between existing ensemble methods and our proposed framework.
Ranker-based ensemble methods, as shown in Figure~\ref{fig:motivation}~(a), aim to select the best candidate from outputs generated by various LLMs~\citep{jiang-etal-2023-llm,tekin-etal-2024-llm}.
However, these methods face several significant limitations: they require quadratic computational cost for pair-wise comparisons, demand additional time for selection or fusion, and encounter scalability issues when scaling to multiple LLMs.
Moreover, ranker-based methods typically operate at the response-level, providing merely coarse-grained ensemble capabilities.

To address these issues, heuristic ensemble methods, shown in Figure~\ref{fig:motivation}~(b), perform ensemble operations at each token generation step~\citep{yu-etal-2024-breaking,huang2024ensemble,mavromatis2024pack,yao2025determinethenensemble}.
While these token-level approaches fuse probability distributions from different models through weighted averaging, they lack consideration of the border context that affects ensemble performance. 
Furthermore, both ranker-based and heuristic ensemble methods face two critical challenges in ensemble weight allocation: (1) they rely on fixed or manually designed rules to assign weights, failing to adapt to domain shifts in input texts (e.g., from mathematical proofs to story writing), thus limiting their generalization ability across diverse contexts, and (2) they inadequately consider context dependencies throughout the generation process, resulting in responses that may be locally optimal but globally suboptimal. Consequently, developing effective LLM ensemble methods remains an open research challenge.

In this paper, we propose a \textit{Reinforcement Learning-assisted Framework for LLM Ensemble} (\method) that reformulates the ensemble process as a reinforcement learning problem.
\method employs RL agent(s) to dynamically adjust the ensemble weights of multiple LLMs based on the input prompt and intermediate generation states, in order to maximize the quality of the final response.
Our framework is grounded in two fundamental insights: (1) different models exhibit context-specific strengths that vary across tasks, and (2) local decisions in reasoning tasks affect global outcomes through path-dependent effects.
Specifically, we formulate the ensemble problem as a Markov Decision Process, where the \textbf{\texttt{state}} contains the input prompt and generated response history, while the \textbf{\texttt{action}} space determines the ensemble weights at each generation step.
To strike a balance between response-level and token-level approaches, \method implements a span-level ensemble strategy, an approach that has shown considerable promise in recent work~\citep{xu-etal-2025-hit,lv2024specfuse}.
While our framework maintains flexibility in its \textbf{\texttt{reward}} function design, we primarily define the output response quality as the reward signal to directly align with the ultimate goal of LLM ensemble.
We instantiate \method using both single-agent and multi-agent reinforcement learning algorithms.
In the multi-agent setting, we treat each LLM as an independent agent while in single-agent setting, one agent is responsible for all the ensemble weights.
Compared to existing LLM ensemble methods, \method achieves an effective balance between ensemble quality and computational efficiency, ensuring the alignment between ensemble objectives and final response quality.

We evaluate our method on a diverse set of benchmarks, including general reasoning, mathematical and scientific problem solving, and code generation.
Experimental results demonstrate that \method outperforms baseline approaches in both performance and computational efficiency. Additionally, our framework exhibits generalization capabilities across different tasks without retraining.

Overall, our \textbf{key contributions} are:
\begin{itemize}
    \setlength{\itemsep}{0pt}
    \item We innovatively formulate the LLM ensemble pipeline as a reinforcement learning task and propose \method, a reinforcement learning-assisted framework for ensemble.
    To the best of our knowledge, this is the first framework that uses reinforcement learning to adaptively optimize ensemble weights.
    \item We employ both single-agent and multi-agent reinforcement learning algorithms to instantiate our framework, each demonstrating distinct advantages in different scenarios.
    \item We provide extensive experimental evidence to demonstrate \method's effectiveness and generalization capabilities, which improves the performance by up to $3.3\%$ accuracy points compared to previous ensemble methods.
\end{itemize}

\begin{figure*}[ht]
    \centering
    \includegraphics[width=\linewidth]{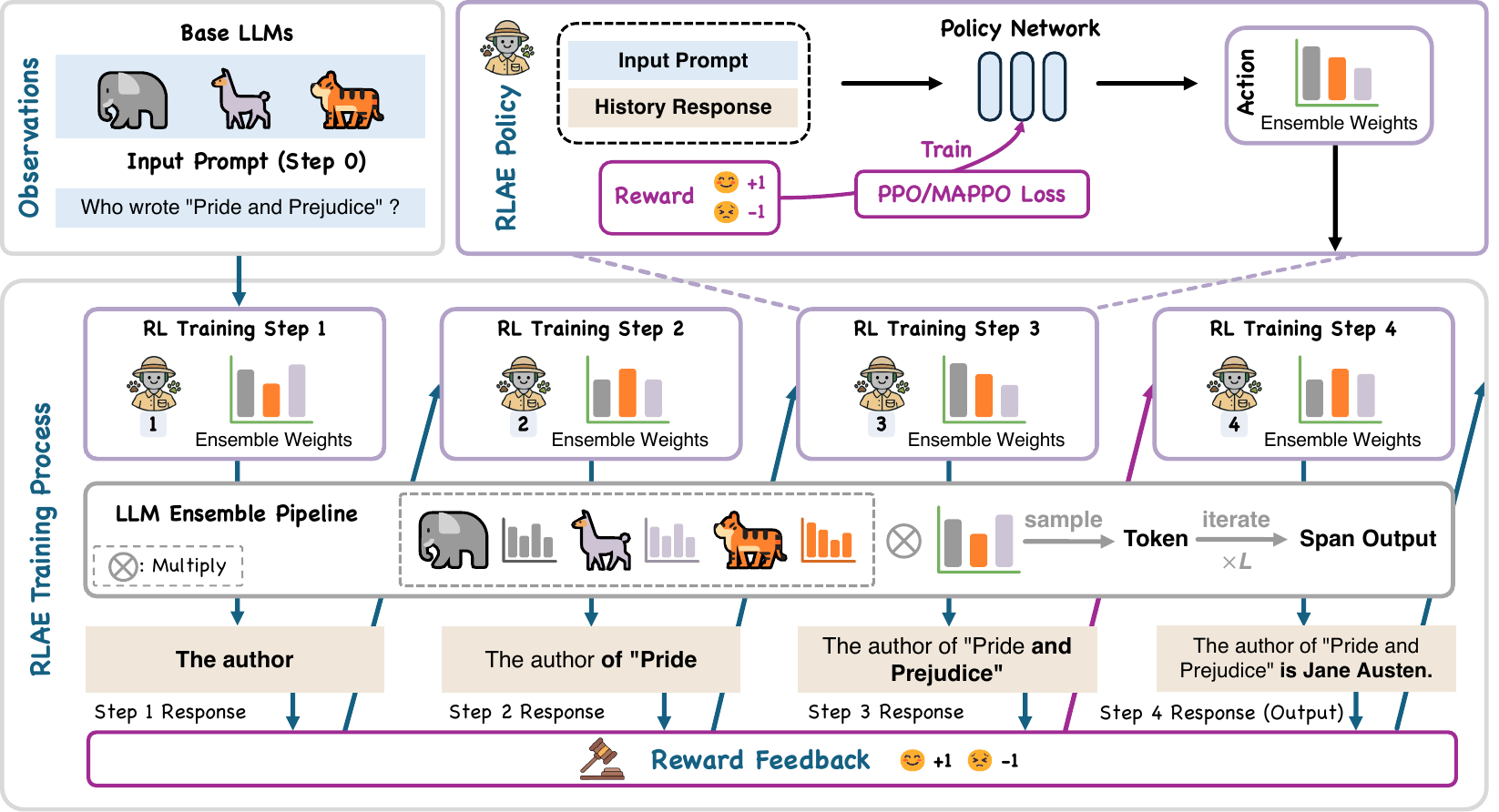}
    \caption{The \method framework. Top row: at each generation step, our RL agent makes actions (ensemble weights) based on the input prompt and history response; the reward feedback is used to train the policy network of the RL agent with the PPO or MAPPO loss. Bottom row: visualization of the iterative generation process of our RL-assisted ensemble framework, where the reward feedback from the final output is continuously used to train the RL agent.}
    \label{fig:framework}
    \vspace{-1em}
\end{figure*}

\section{Related Works}
\paragraph{LLM Ensemble.}
Recent advances in LLM capabilities have sparked significant interest in ensemble methods that combine multiple models to achieve superior performance.
Current approaches can be categorized into three main types: token-level, response-level, and span-level ensembles.
Token-level methods, such as \textsc{GaC}~\citep{yu-etal-2024-breaking}, \textsc{DeePEn}~\citep{huang2024ensemble}, and \textsc{PackLLM}~\citep{mavromatis2024pack}, combine probability distributions during generation through vocabulary alignment and distribution projection. While these methods enable fine-grained ensemble, they often compromise semantic coherence and struggle with computational complexity.
Response-level approaches, including \textsc{LLM-Blender}~\citep{jiang-etal-2023-llm} and \textsc{LLM-TOPLA}~\citep{tekin-etal-2024-llm}, select or combine complete responses after generation. 
While these methods are practical for black-box LLMs, they are unable to effectively leverage the complementary strengths of different models.
Besides, both approaches are constrained by fixed weight allocation strategies and insufficient consideration of context dependencies.
In this paper, we propose a RL-assisted framework that dynamically adjusts ensemble weights based on the prompt and generation history.
Our method aligns with span-level approaches, such as \textsc{SweetSpan}~\citep{xu-etal-2025-hit} and \textsc{SpecFuse}~\citep{lv2024specfuse}, which operate on sequences of spans to achieve a balance between granularity and efficiency.

\paragraph{Reinforcement Learning in LLM.}
Reinforcement learning has been extensively used to enhance LLM capabilities. Reinforcement learning from human feedback (RLHF) leverages human preferences to guide model outputs~\citep{ouyang2022training,tu2025online}, while reinforcement fine-tuning (RFT) employs task-specific rewards for performance improvements in tasks~\citep{guo2025deepseek,xu2025diplollm,chai2025empowering,yu2025dapo}. Additionally, LLM routing~\citep{zheng2025citer} or mixture-of-agents~(MoA)~\citep{chakraborty2025collab} methods employ RL to dynamically select the most appropriate LLM or expert based on input prompt or intermediate generation steps. However, these applications typically focus on selecting or fine-tuning individual models rather than coordinating multiple models for ensemble. In contrast, our framework represents the first to apply RL for adaptive ensemble weight adjustment, enabling the integration of the strengths of different models.

\section{Methodology}

In this section, we introduce \method, a reinforcement learning-assisted framework designed for dynamic LLM ensemble that facilitates collaborative inference among models with complementary strengths. As illustrated in Figure~\ref{fig:framework}, \method employs a RL agent to adjust ensemble weights based on both the input prompt and intermediate generation responses.
We organize this section as follows: Section~\ref{sec:prob_formulation} formalizes the LLM ensemble problem, Section~\ref{sec:rl_formulation} details our \textit{Reinforcement Learning-Assisted LLM Ensemble Framework}, and Section~\ref{sec:training_methodology} presents our training methodology.

\subsection{Problem Formulation}
\label{sec:prob_formulation}

\paragraph{LLM Ensemble Problem.}
Consider a set of $K$ foundational LLMs $\mathcal{M} = \{M_1, \ldots, M_K\}$ with corresponding parameters $\{\theta_1, \ldots, \theta_K\}$. For a given input prompt $\bm{x} = (x_1, \ldots, x_m)$ of length $m$, the ensemble generates an output sequence $\bm{y} = (y_1, \ldots, y_H)$ of length $H$ through iterative probability fusion:
\begin{equation}
    \label{eq:ensemble_prob}
p(y_h|\bm{x}, \bm{y}_{<h}) = \sum_{k=1}^K w_h^{(k)} p_{M_k}(y_h|\bm{x}, \bm{y}_{<h}, \theta_k),
\end{equation}
where $w_h^{(k)} \in [0,1]$ denotes the ensemble weight assigned to model $M_k$ at generation step $h$, satisfying the constraint $\sum_{k=1}^K w_h^{(k)} = 1$, and $p_{M_k}(y_h|\cdot)$ is the probability of the $h$-th token generated by model $M_k$. The history context $\bm{c}_h = (\bm{x}, \bm{y}_{<h})$ contains both the input prompt and previously generated tokens, while a weight function $\pi_\phi: \mathcal{C} \times \mathcal{M} \to \Delta^{K-1}$, parameterized by $\phi$, determines the weights over the $(K-1)$-dimensional probability simplex.

\paragraph{Span-Level Ensemble.}

Traditional LLM ensemble approaches~\citep{yu-etal-2024-breaking,huang2024ensemble,mavromatis2024pack} typically combine probabilities across the entire vocabulary at each generation step. This approach imposes significant computational overhead during inference, adversely affecting both performance and efficiency.
To address these challenges while maintaining semantic coherence, we implement span-level ensemble, an approach increasingly explored in recent work~\citep{xu-etal-2025-hit,lv2024specfuse}. Instead of adjusting weights for individual tokens, our method applies consistent weights at the span level, where each span $\bm{z}_t$ comprises $L$ consecutive tokens:
\begin{equation}
\bm{z}_t = (y_{t,1}, y_{t,2}, \ldots, y_{t,L}),
\end{equation}
where $\bm{z}_t$ represents tokens generated in the $t$-th span, and $L$ is a predefined span length.
This strategic approach reduces decision points from $H$ (token-level) to $\lceil H/L \rceil$ (span-level), substantially improving computational efficiency while ensuring semantic continuity. 
This design enables the RL agent to focus on examining model contributions at span-level, streamlining the state-action space for RL training and enhancing response quality. 
Furthermore, following \textsc{GaC}~\citep{yu-etal-2024-breaking}, we recognize that not all generation steps in the span necessarily require ensembling. 
Therefore, we selectively ensemble only critical tokens within each span, further optimizing computational efficiency without compromising generation quality.

\paragraph{Essential Elements of MDP.}

We formulate the LLM ensemble as a Markov Decision Process (MDP) with the following components:
\begin{itemize}
    \setlength{\itemsep}{0pt}
    \item \textbf{\texttt{State}} ($s_t$): The state is represented as $s_t=(\bm{x},\bm{z}_{<t})$, capturing both the input prompt and the response up to the current span $t$.
    
    \item \textbf{\texttt{Action}} ($a_t$): The action at span $t$ is defined as $a_t=(w_t^{(1)},\ldots,w_t^{(K)})$, representing the ensemble weights across models for the next span generation, where $\sum_{k=1}^K w_t^{(k)} = 1$ and $w_t^{(k)} \in [0,1]$ for all $k$.
    
    \item \textbf{\texttt{Reward}} ($r(s_t,a_t)$): The reward function evaluates the generation quality. For terminal states, it uses task-specific metrics. For intermediate states, it employs a process reward model that provides dense feedback.
    
    \item \textbf{\texttt{Transition}} ($P(s_{t+1}|s_t,a_t)$):  
    The transition function defines the environment dynamics based on the current state and action.
    In LLM ensemble, this is determined by the weighted combination of model probabilities for generating the next span of tokens.
    
    \item \textbf{\texttt{Policy}} ($\pi_\phi(a_t|s_t)$): As the weight function mentioned before, the policy function defines the ensemble weight distribution across models based on the current state.
\end{itemize}

Previous LLM ensemble approaches often compute fixed weights across the entire dataset~\citep{chen2025harnessing,lu2024merge}, which fails to capture the context-dependent nature of model performance. These methods typically assign fixed weights to each model based on performance metrics or heuristic rules, ignoring how model strengths vary across different inputs and generation steps. In contrast, our reinforcement learning-assisted framework, \method, learns to dynamically adjust weights based on the prompts and intermediate generation states, effectively adjusting the ensemble weights.

\subsection{Reinforcement Learning-Assisted LLM Ensemble Framework (\method)}
\label{sec:rl_formulation}

Our ensemble procedure operates in an autoregressive manner, which adjusts weights dynamically at the end of each span.
The process is formalized in Algorithm~\ref{alg:ensemble} in Appendix~\ref{sec:pseudocode}.
The algorithm begins with an empty output sequence and iteratively builds the response span by span. 
At each span, the RL agent evaluates the current state, which includes the input prompt and generated responses. Based on this evaluation, the RL agent samples a weight distribution across the base models, determining their relative contributions to the next span. These weights remain fixed throughout the span generation, with each token sampled from the weighted probability distribution of the ensemble. This process continues until reaching either the desired output length or the termination token \texttt{<EOS>}.
The RL agent's policy $\pi_\phi$ learns to assign appropriate weights by analyzing input patterns and partial generations, effectively identifying which models are most reliable for different contexts. This dynamic adaptation allows the ensemble to leverage the strengths of each base model while mitigating their individual weaknesses, resulting in higher quality generations.
Another key challenge in LLM ensemble stems from the fact that different LLMs are trained based on different tokenizers, resulting in diverse vocabulary sets across models.
To address vocabulary mismatch, \method projects probability vectors from multiple LLMs into a unified vocabulary space through a mapping matrix. More details can be found in Appendix~\ref{sec:vocabulary_mismatch}.

\subsection{Training Process of \method}
\label{sec:training_methodology}

For optimization, the RL agent employs Proximal Policy Optimization (PPO)~\citep{schulman2017proximal} for single-agent training and its multi-agent variant (MAPPO)~\citep{yu2022surprising} for multi-agent scenarios.
The training process consists of iteratively collecting trajectories by executing the current policy on input prompts, followed by policy and value function optimization. 
Algorithm~\ref{alg:training} in Appendix~\ref{sec:pseudocode} provides a detailed description of this process.

\subsubsection{Single-Agent Training}

In the single-agent version \methodppo, we formulate the ensemble process as a centralized decision-making problem, where a single RL agent controls the weights of all LLMs simultaneously.
At each step, the agent observes the current state and outputs a weight distribution over base models, determining their relative contributions to the output. 
We optimize \methodppo using PPO, which employs a policy network (actor) that dynamically adjusts ensemble weights, coupled with a value network (critic) that estimates expected returns. 
The objective function of PPO is:
\begin{equation}
    \label{eq:ppo_objective}
    \begin{aligned}
    L&_\text{PPO}(\phi)=\min\left(\frac{\pi_\phi(a|s)}{\pi_{\phi_{old}}(a|s)}A^{\pi_{\phi_{old}}}(s,a),\right.\\
    &\left.\mathrm{clip}\left(\frac{\pi_\phi(a|s)}{\pi_{\phi_{old}}(a|s)},1-\epsilon,1+\epsilon\right)A^{\pi_{\phi_{old}}}(s,a)\right),
    \end{aligned}
\end{equation}
where $\pi_{\phi}/\pi_{\phi_{old}}$ denotes the importance sampling ratio, 
which measures the difference between the new and old policies, and $\epsilon$ serves as the clipping parameter constraining policy updates.
The advantage function $A(s,a)$ is estimated using GAE~\citep{schulman2015high}:
\begin{equation}
    A(s,a)=\sum_{l=0}^{\infty}\left(\gamma\lambda\right)^l \delta_{t+l},
\end{equation}
\begin{equation}
    \delta_{t}=r_{t}+\gamma V(s_{t+1})-V(s_{t}),
\end{equation}
where $\gamma$ is the discount factor, $\lambda$ is the GAE parameter, $\delta_{t}$ is the temporal difference (TD) error, and $V(s)$ is the value function.
The value network is trained to minimize the mean squared error between its predictions and the actual returns, which is detailed in Appendix~\ref{sec:value_network}.

\subsubsection{Multi-Agent Training}

As the number of base models increases, the single-agent approach faces challenges in managing the expanding joint action space and modeling complex interactions between base models. 
To address these limitations, we reformulate the problem as a Markov game (see Appendix~\ref{sec:markov_game} for details) where each LLM is controlled by an independent RL agent that determines its model's contribution weight.
This design offers three advantages: (1) It decomposes the complex joint action space into smaller individual action spaces, reducing the dimensionality of the learning problem for each agent; (2) It enables each RL agent to specialize in understanding its corresponding model's strengths and weaknesses; and (3) It facilitates more flexible scaling through independent agent addition.

In \methodmappo, we optimize RL agents using MAPPO, which extends PPO to a multi-agent setting. Each RL agent outputs a scalar logit value through its own policy network. These logits are then concatenated and passed through a \texttt{Softmax} function to obtain the final ensemble weights that sum to 1. The agents share a centralized critic that coordinates global rewards, following the centralized training with decentralized execution (CTDE) paradigm~\citep{gronauer2022multi}.
The objective function of MAPPO is:
\begin{equation}
    \label{eq:mappo_objective}
    \begin{aligned}
    L&_\text{MAPPO}(\phi)=\frac{1}{K}\sum_{k=1}^K\min\left(\frac{\pi_\phi(a^{(k)}|s)}{\pi_{\phi_{old}}(a^{(k)}|s)}A^{(k)},\right.\\
    &\left.\mathrm{clip}\left(\frac{\pi_\phi(a^{(k)}|s)}{\pi_{\phi_{old}}(a^{(k)}|s)},1-\epsilon,1+\epsilon\right)A^{(k)}\right),
    \end{aligned}
    \end{equation}
where $\pi_{\phi}(a^{(k)}|s)$ is the policy of agent $k$.
The centralized critic takes the global state as input to better estimate the value function, facilitating more effective credit assignment across agents.
This design allows for context-aware weight adjustments while maintaining alignment with overall ensemble objectives through shared reward signals.

\section{Experiments}

\subsection{Experimental Settings}

\paragraph{Benchmarks.}

We evaluate our method on seven benchmarks across three capability dimensions: (1) \textbf{General Ability}: MMLU (0-shot)~\citep{hendrycks2021measuring}, a multiple-choice dataset covering 57 subjects across STEM, humanities, and social sciences; ARC-C (0-shot)~\citep{clark2018think}, containing questions from standardized science exams for grades 3-9;  and TriviaQA (5-shot)~\citep{joshi-etal-2017-triviaqa}, a factual question-answering dataset compiled by trivia enthusiasts to test retrieval of world knowledge. (2) \textbf{Math and Science Ability}: GSM8K (5-shot)~\citep{cobbe2021gsm8k}, featuring linguistically diverse grade school math word problems requiring multi-step reasoning; PIQA (0-shot)~\citep{Bisk2020}, a physical commonsense reasoning dataset; and GPQA (5-shot)~\citep{rein2024gpqa}, a graduate-level professional question-answering benchmark focusing on physics, chemistry, and biology. (3) \textbf{Code Generation}: MBPP (0-shot)~\citep{austin2021program}, where models generate Python code for basic programming problems designed for entry-level programmers.

\paragraph{Base Models.}
As ensemble methods typically work better with models of comparable performance but diverse capabilities, we select a set of models with similar parameter scales. Our base LLMs includes Llama-3.1-8B-Instruct~\citep{grattafiori2024llama3}, Qwen-2-7B-Instruct~\citep{yang2024qwen2}, Qwen-2.5-7B-Instruct~\citep{qwen2025qwen25}, Mistral-7B-Instruct-v0.3~\citep{jiang2023mistral}, and OpenChat-3.5~\citep{wang2024openchat}.

\paragraph{Baselines.}
We compare our method against three representative LLM ensemble approaches: (1) \textsc{\textbf{PairRanker}}, the key component of \textsc{LLM-Blender}~\citep{jiang-etal-2023-llm}, which employs pairwise comparison to evaluate candidate outputs from different LLMs and selects the highest-scoring response as the final output (we omit \textsc{GenFuser} due to its significant refusal rate on our benchmarks);
(2) \textsc{\textbf{GaC}}~\citep{yu-etal-2024-breaking}, which constructs a union dictionary by combining vocabularies from multiple models and projects each model's token distribution onto this unified space for aggregated token selection at each generation step; 
and (3) \textsc{\textbf{DeePEn}}~\citep{huang2024ensemble}, which leverages relative representation theory to map probability distributions from different models into a shared space for aggregation, using the intersection of model vocabularies as the basis for this projection. These baselines represent diverse approaches to LLM ensemble, spanning ranker-based ensemble and heuristic ensemble methods.

\paragraph{Implementation Details.}

For the RL agent architecture, we select DeBERTa-V3-Large~\citep{he2021debertav3}, a compact yet powerful model with around 400M parameters to minimize training overhead. 
For reward design, we employ a rule-based sparse reward model that provides feedback only at the terminal state of generation, utilizing benchmark-specific evaluation metrics such as accuracy. Although reward computation introduces some computational overhead during training, it is worth noting that \method operates without requiring any reward or supervision signals at inference time.

\addtolength{\extrarowheight}{\belowrulesep}
\aboverulesep=0pt
\belowrulesep=0pt

\begin{table*}[ht]
    \renewcommand{\arraystretch}{1.1}
    \centering
    \resizebox{\textwidth}{!}{%
    \begin{tabular}{>{}lcccccccccc<{}}
    \toprule
    \multicolumn{1}{c}{\multirow{2}{*}{\textbf{Dataset}}} &
      \multicolumn{2}{c}{\textbf{MMLU}} &
      \multicolumn{2}{c}{\textbf{ARC-C}} &
      \multicolumn{2}{c}{\textbf{GPQA}} &
      \multicolumn{2}{c}{\textbf{GSM8K}} &
      \multicolumn{2}{c}{\textbf{MBPP}}  \\ \cmidrule(l){2-11}
    \multicolumn{1}{c}{} &
      \multicolumn{1}{l}{2 LLMs} &
      \multicolumn{1}{l}{3 LLMs} &
      \multicolumn{1}{l}{2 LLMs} &
      \multicolumn{1}{l}{3 LLMs} &
      \multicolumn{1}{l}{2 LLMs} &
      \multicolumn{1}{l}{3 LLMs} &
      \multicolumn{1}{l}{2 LLMs} &
      \multicolumn{1}{l}{3 LLMs} &
      \multicolumn{1}{l}{2 LLMs} &
      \multicolumn{1}{l}{3 LLMs} 
       \\ \midrule
    Llama-3.1-8B-Instruct & 66.8 & 66.8 & 79.5 & 79.5 & 32.8 & 32.8 & 84.5 & 84.5 & 69.6 & 69.6     \\
    Qwen-2-7B-Instruct    & 65.3 & 65.3 & 81.9 & 81.9 & 34.3 & 34.3 & 85.7 & 85.7 & 67.2 & 67.2     \\
    Qwen-2.5-7B-Instruct  & 68.2 & 68.2 & 84.3 & 84.3 & 36.4 & 36.4 & 91.6 & 91.6 & 76.3 & 76.3     \\ \midrule
    \textsc{PairRanker}~\citep{jiang-etal-2023-llm}            & 63.8 & 69.1 & 78.6 & 82.8 & 32.7 & 34.2 & 86.8 & \textbf{91.7} & 66.1 & 64.5  \\
    \textsc{GaC}~\citep{yu-etal-2024-breaking}                   & 67.5 & 67.8 & 82.1 & 84.5 & \textbf{35.0} & 33.4 & 83.1 & 88.1 & 68.6 & 74.6  \\
    \textsc{DeePEn}~\citep{huang2024ensemble}                & 67.1 & 68.0 & 81.8 & 84.1 & 33.8 & 32.6 & 84.2 & 86.2 & \textbf{69.9} & 69.9  \\ \hdashline
    \textbf{\methodppo}            & \textbf{68.5} & \textbf{69.2} & \textbf{82.8} & \textbf{84.7} & \cellcolor[HTML]{BFFFFF}\textbf{35.1} & \cellcolor[HTML]{BFFFFF}\textbf{36.1} & \textbf{86.9} & 91.3 & \cellcolor[HTML]{BFFFFF}\textbf{70.5} & \cellcolor[HTML]{BFFFFF}\textbf{75.8}  \\
    \textbf{\methodmappo}  & \cellcolor[HTML]{BFFFFF}\textbf{69.1} & \cellcolor[HTML]{BFFFFF}\textbf{70.1} & \cellcolor[HTML]{BFFFFF}\textbf{83.4} & \cellcolor[HTML]{BFFFFF}\textbf{85.6} & 34.7 & \textbf{35.3} & \cellcolor[HTML]{BFFFFF}\textbf{87.4} & \cellcolor[HTML]{BFFFFF}\textbf{92.5} & 69.8 & \textbf{75.3} \\ \bottomrule
    \end{tabular}%
    }
    \caption{Performance of LLM ensemble. 2 LLMs mean Llama-3.1 and Qwen-2; 3 LLMs mean Llama-3.1, Qwen-2, and Qwen-2.5. \colorbox[HTML]{BFFFFF}{\textbf{Highlight}} indicates the best performing method, while \textbf{bold} indicates the second best.}
    \label{tab:main_results}
    \end{table*}
\addtolength{\extrarowheight}{\belowrulesep}
\aboverulesep=0pt
\belowrulesep=0pt
\begin{table}[ht]
    \centering
    \renewcommand{\arraystretch}{1}
    \resizebox{\columnwidth}{!}{%
    \begin{tabular}{>{}lccccc<{}}
    \toprule
    \multicolumn{1}{c}{\multirow{2}{*}{\textbf{Model}}} &
      \multicolumn{5}{c}{\textbf{Dataset}}  \\ \cmidrule(lr){2-6}
    \multicolumn{1}{c}{} &
      \textbf{MMLU} &
      \textbf{ARC-C} &
      \textbf{TriviaQA} &
      \textbf{GSM8K} &
      \textbf{PIQA} 
       \\ \midrule 
    Mistral-7B-Instruct-v0.3 & 59.3 & 74.5 & 64.3 & 56.5 & 80.6  \\
    OpenChat-3.5             & 60.8 & 78.1 & 61.7 & 73.4 & 87.1  \\ \midrule
    \textsc{PairRanker}               & 60.3 & 75.9 & 56.3 & 67.7 & \cellcolor[HTML]{BFFFFF}\textbf{82.9}  \\
    \textsc{GaC}                      & 55.3 & 73.8 & 62.2 & 60.8 & 69.2 \\
    \textsc{DeePEn}                   & \textbf{61.8} & 71.6 & \textbf{67.3} & \cellcolor[HTML]{BFFFFF}\textbf{69.4} & 77.5  \\ \hdashline
    \textbf{\methodppo}               & 61.1 & \cellcolor[HTML]{BFFFFF}\textbf{79.2} & \cellcolor[HTML]{BFFFFF}\textbf{67.5} & 66.6 & \textbf{81.2}  \\
    \textbf{\methodmappo}             & \cellcolor[HTML]{BFFFFF}\textbf{62.5} & \textbf{78.4} & 65.6 & \textbf{67.9} & 80.8 \\ \bottomrule
    \end{tabular}
    }
    \caption{Performance of ensemble methods with Mistral 7B and OpenChat 3.5.}
    \label{tab:main_results_2}
\end{table}

\subsection{Main Results}

Tables~\ref{tab:main_results} and~\ref{tab:main_results_2} demonstrate the performance of our proposed \method framework compared to base models and baseline ensemble methods across various benchmarks. Overall, our method consistently outperforms existing methods in most scenarios, with \methodmappo achieving a superior average score of $70.1$ in the primary experimental set, surpassing all base models and ensemble methods. From these results, we derive several observations:

\paragraph{(1) Effective Performance Enhancement with Comparable Base Models.} Our experiments demonstrate significant improvements when ensembling models with similar performance levels. For instance, when combining Llama-3.1 and Qwen-2, which have a relatively small performance gap of $1.5$ points on MMLU, \methodmappo achieves a substantial $2.3$-point improvement over the stronger base model. We observe similar gains on ARC-C, where \methodmappo increases performance to $83.4$ points, representing a $1.5$-point improvement over base model performance. However, the improvements become more modest when incorporating Qwen-2.5, which outperforms other models by $2.9$ points on MMLU. In this case, the ensemble yields only a $1.9$-point improvement to reach $70.1$ points, highlighting the challenges in effectively combining models with larger performance disparities.

\paragraph{(2) \methodmappo Superiority with Heterogeneous Base Models.} In scenarios with significant performance variations among base models, \methodmappo demonstrates advantages over \methodppo. This superiority stems from several key factors: First, the multi-agent framework allows each agent to specialize in modeling a specific LLM's behavior patterns and output characteristics, leading to more accurate weight assignments. Second, the shared critic in MAPPO enables agents to learn from each other's experiences while maintaining their individual policies, facilitating better coordination when base models have complementary strengths.
For example, we observe that \methodmappo learns to leverage Qwen-2's superior performance on STEM questions, which are shown in Appendix~\ref{sec:visualization_ensemble_weights}. 
These advantages are empirically validated in our three-model ensemble results, where \methodmappo achieves a $0.3$ point improvement over the single-agent \methodppo, with particularly high gains on questions requiring diverse domain expertise.

\paragraph{(3) \methodppo Advantage in Code Generation.} Notably, \methodppo outperforms \methodmappo on the MBPP programming benchmark ($75.8$ versus $75.3$ in three-model settings). We attribute this to the nature of code generation tasks, which require maintaining global consistency throughout the generation process. 
This aligns with findings from traditional reinforcement learning research, where single-agent approaches provide more consistent control and better performance in tasks requiring coordinated actions, as demonstrated in robotics control tasks like MuJoCo simulations~\citep{brockman2016openai,peng2021facmac}.

\subsection{Analysis}

To comprehensively evaluate \method, we conduct extensive analyses focusing on three critical aspects: computational efficiency in terms of time latency, generalization capabilities across different tasks, and ablation studies on key components. All analysis experiments utilize a two-LLM ensemble configuration (Qwen-2 and LLaMA-3.1).

\begin{figure}[ht]
    \centering
    \includegraphics[width=\linewidth]{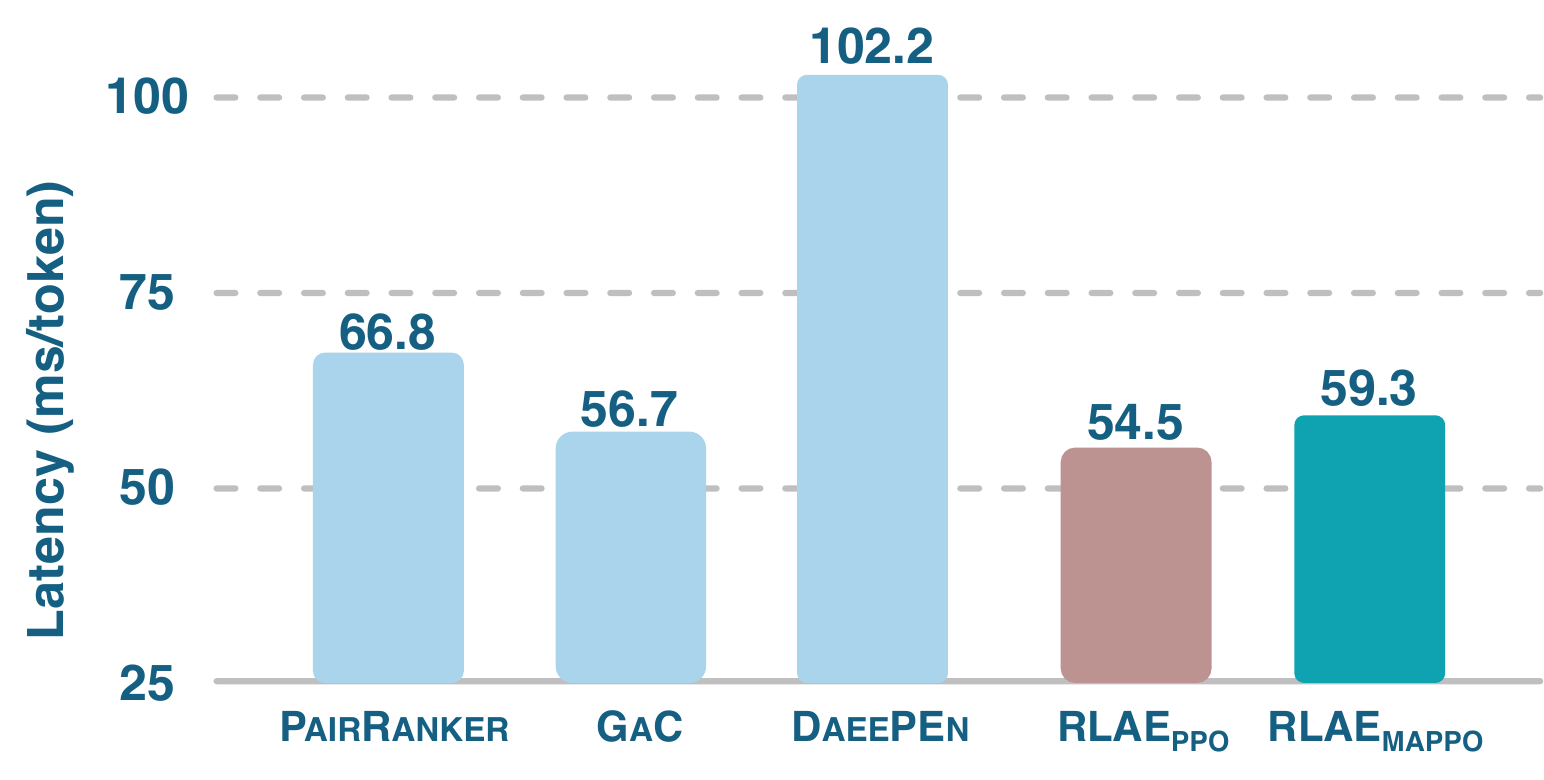}
    \caption{Time latency comparison of different methods.}
    \label{fig:latency}
\end{figure}

\paragraph{Latency Analysis.}
In order to evaluate the computational efficiency of our method, we test the latency (ms/token) across different ensemble configurations. As illustrated in Figure~\ref{fig:latency}, our method achieves latency comparable to \textsc{GaC}, which is substantially lower than that of \textsc{PairRanker} and \textsc{DeePEn}. 
Despite incorporating an additional RL agent with 400M parameters, our method maintains competitive efficiency through span-level ensemble optimization, which reduces the frequency of weight adjustments.

\begin{table}[ht]
    \centering
    \renewcommand{\arraystretch}{1.1}
    \resizebox{\columnwidth}{!}{%
    \begin{tabular}{@{}cccc@{}}
    \toprule
    \textbf{Dataset}    & \textbf{\textsc{PairRanker}}  & \textbf{\methodppo}   & \textbf{\methodmappo}   \\ \midrule
    \textbf{ARC-C}      & 78.6        & 82.8        & 83.4        \\
    \textbf{MMLU$\,\to\,$ARC-C} & 74.8 (-3.8) & \textbf{82.2 (-0.6)} & \textbf{83.0 (-0.4)} \\ \bottomrule
    \end{tabular}
    }
    \caption{Generalization of \method and \textsc{PairRanker} from MMLU to ARC-C.}
    \label{tab:generalize}
    \end{table}

\paragraph{Generalization of RL Agent.}
To assess the generalization capability of our method, we conduct a cross-task evaluation by directly applying the RL agent and \textsc{PairRanker} trained on MMLU to ensemble LLMs on ARC-C, a related but distinct task. As shown in Table~\ref{tab:generalize}, while \textsc{PairRanker} exhibits significant performance degradation on ARC-C ($3.8$ points lower compared to direct training), our method demonstrates remarkable generalization with only minimal performance drops ($0.6$ and $0.4$ points for PPO and MAPPO, respectively). These results highlight the  generalization capabilities of \method across different tasks. More results can be found in Appendix~\ref{sec:add_results}.

\paragraph{Effect of Ensemble Weights by RL.}

To evaluate the effectiveness of our RL-based weight generation approach, we conduct comparative experiments against baseline weighting strategies, including uniform weighting and perplexity-based weighting. For perplexity-based weighting, we calculate the perplexity (PPL) score as:
\begin{equation}
    \text{PPL}_k(\bm{x}) = \exp\left(-\frac{1}{m}\sum_{i=1}^{m}\log p_{M_k}(x_i|x_{<i})\right)
\end{equation}
where lower PPL indicates higher model confidence and receives the higher ensemble weight.

\begin{figure}[htb]
    \centering
    \includegraphics[width=\linewidth]{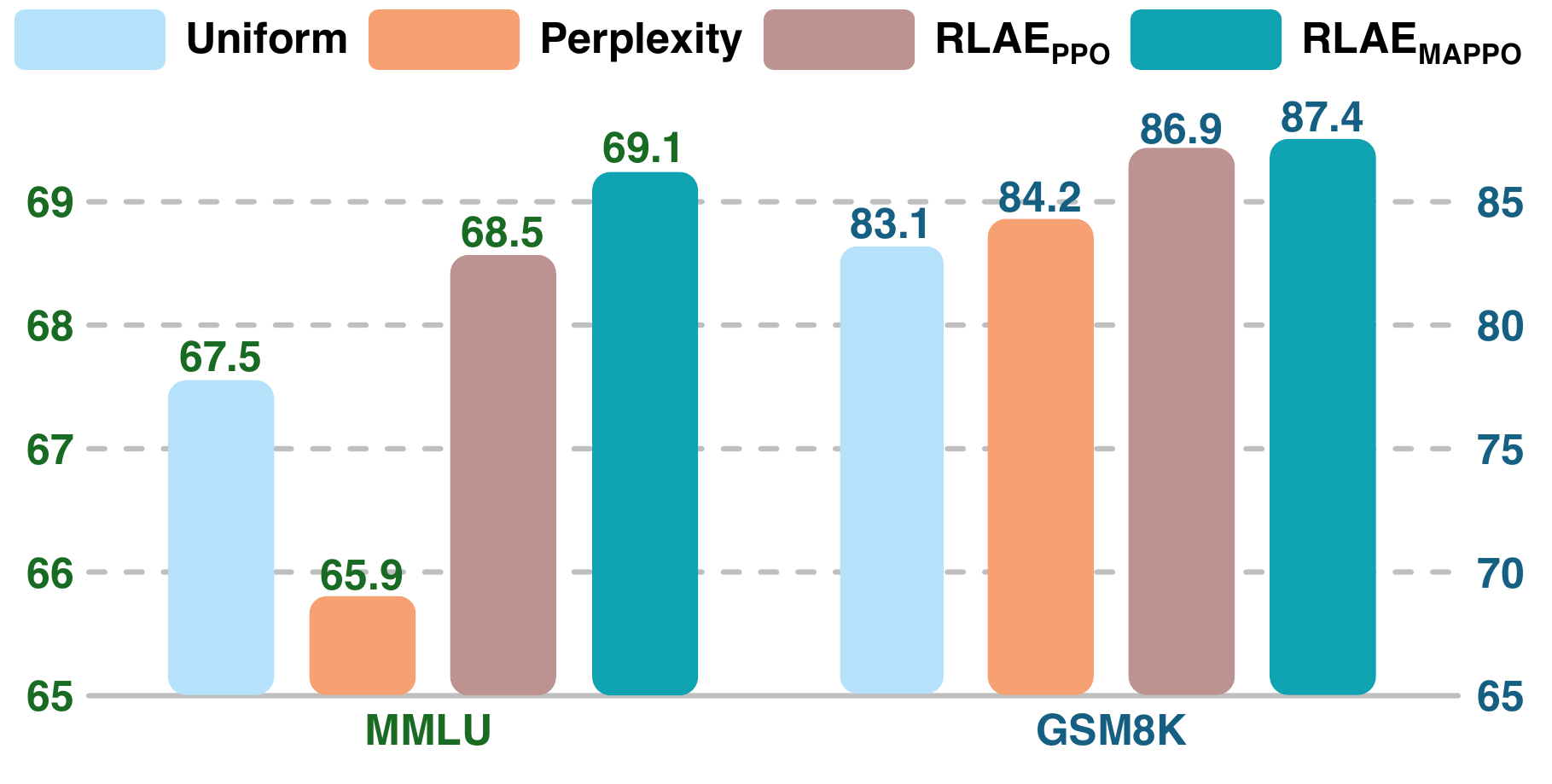}
    \caption{Performance comparison of different weighting methods across tasks.}
    \label{fig:weight}
    \vspace{-0.25em}
\end{figure}

As shown in Figure~\ref{fig:weight}, \method outperforms baselines in both single-agent and multi-agent settings. 
Besides, uniform weighting performs better on MMLU, while perplexity-based weighting achieves superior results on GSM8K. 
This suggests that previous methods require task-specific weight adjustments, whereas \method automatically adapts to different tasks without manual intervention.

\begin{figure}[ht]
    \centering
    \includegraphics[width=\linewidth]{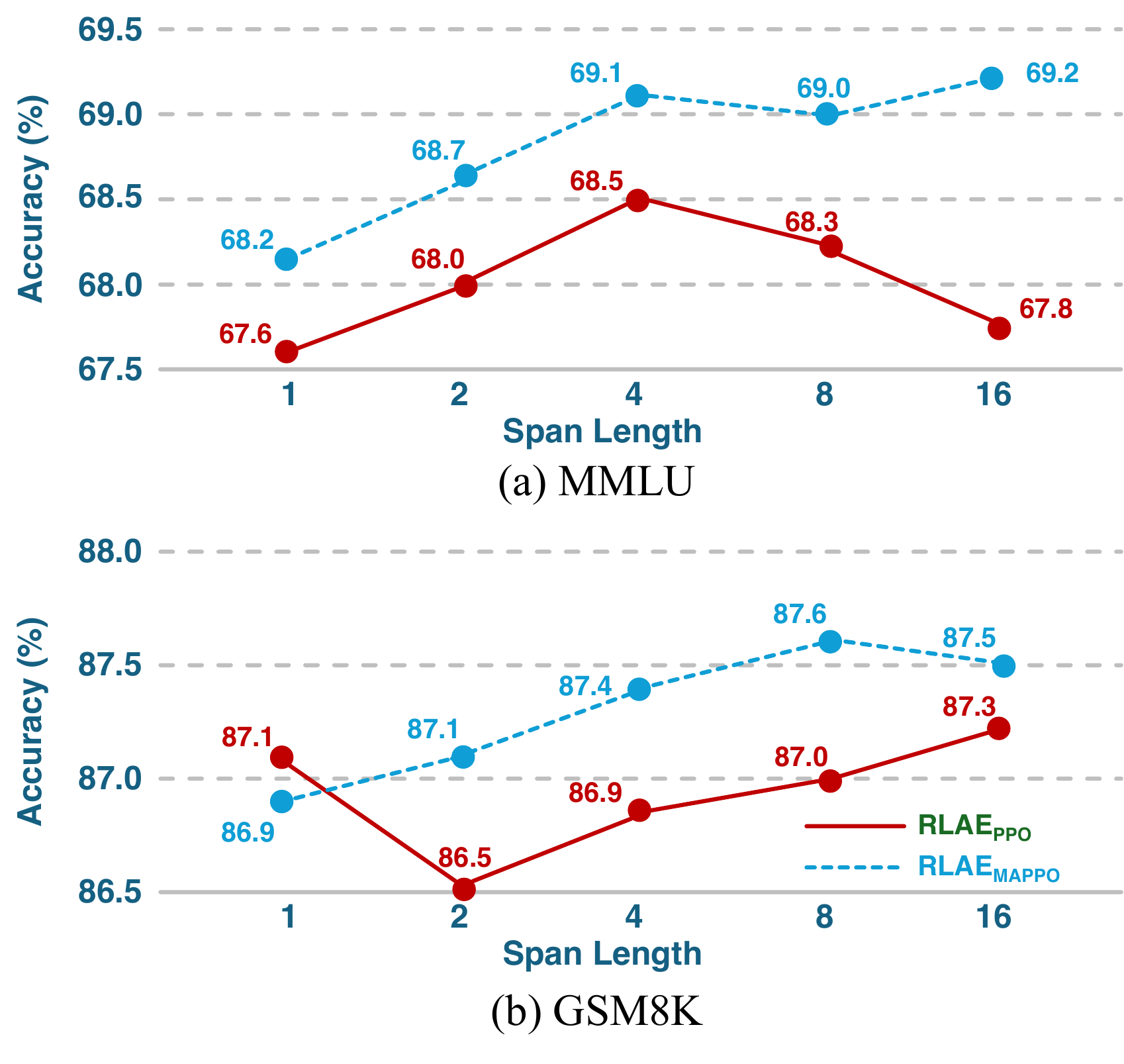}
    \vspace{-1.5em}
    \caption{Span length ablation study on different tasks.}
    \label{fig:span}
    \vspace{-0.5em}
\end{figure}

\paragraph{Effect of the Span Length.}

To explore the impact of span length on performance, we conduct an ablation study varying the span length from 1 to 16 and evaluate the performance of different methods. The ablation results are shown in Figure~\ref{fig:span}, indicating that \methodppo is more sensitive to span length compared to \methodmappo. Additionally, the impact of span length on performance varies across different tasks. 
The span length of $4$ used in our main experiments proves to be a balanced choice.

\section{Conclusion}
In this paper, we introduce \method, a novel reinforcement learning-assisted framework for LLM ensemble that significantly enhances LLM capabilities by dynamically combining the complementary strengths of different models.
By formulating the ensemble problem as a Markov Decision Process at the span level, \method adaptively adjusts ensemble weights based on both prompt and responses, enabling flexible and efficient generation.
Unlike previous ensemble methods, we employ single-agent and multi-agent RL algorithms to optimize the ensemble process.
Extensive experiments demonstrate that our RL-based framework achieves not only substantial improvements but also generalization capabilities across different tasks.

\clearpage

\section*{Limitations}
Due to the inherent nature of the ensemble, our approach, like other ensemble methods, requires increased computational resources compared to single-model inference. While parallel execution on separate GPUs limits latency to that of the slowest model, the computational demand still scales linearly with the number of models. This creates resource challenges, especially during initial inference. 
Furthermore, the effectiveness of our method is closely tied to reward design, where current metrics such as accuracy may not comprehensively align with generation quality.

Future work could explore efficient model selection mechanisms that identify an optimal subset of models prior to inference, thereby reducing computational overhead while maintaining ensemble effectiveness.
Additionally, other promising research directions include developing reward modeling approaches that better capture generation quality, and incorporating human feedback to improve alignment with human preferences.

\bibliography{custom,anthology}

\clearpage
\appendix

\renewcommand{\thefigure}{A\arabic{figure}}
\renewcommand{\thetable}{A\arabic{table}}

\setcounter{figure}{0}
\setcounter{table}{0}

\renewcommand{\theequation}{A\arabic{equation}}
\setcounter{equation}{0}

\begin{tcolorbox}[colframe=black!80, colback=white, sharp corners, boxrule=1pt, arc=5pt, rounded corners, left=1pt, right=1pt, top=2pt, bottom=2pt]
    \centering
    \large \textbf{Appendix}
\end{tcolorbox}

\section{Pseudocode of \method}
\label{sec:pseudocode}
We describe the ensemble generation and training process in Algorithm~\ref{alg:ensemble} and Algorithm~\ref{alg:training}.

\begin{algorithm}[h!]
    \renewcommand{\arraystretch}{1}
    \caption{Ensemble Generation for \method}
    \label{alg:ensemble}
    \begin{algorithmic}
    \STATE \textbf{Input:} Prompt $\bm{x}$, base models $\{M_1, \ldots, M_K\}$, controller policy $\pi_\phi$, span length $L$
    \STATE \textbf{Output:} Generated response $\bm{y}$
    \STATE Initialize $\bm{y}_0 \leftarrow \emptyset$
    \FOR{$t = 0, 1, \ldots, \lceil H/L \rceil - 1$}
        \STATE \textcolor[HTML]{0671b9}{ \texttt{// Construct current state}}
        \STATE $s_t \leftarrow (\bm{x}, \bm{z}_{<t})$
        \STATE \textcolor[HTML]{0671b9}{ \texttt{// Sample weights from policy}}
        \STATE $a_t \leftarrow (w_t^{(1)}, \ldots, w_t^{(K)}) \sim \pi_\phi(\cdot|s_t)$
        \FOR{$h = tL, \ldots, \min\left((t+1)L-1, H\right)$}
            \STATE Generate ensemble probability by Eq.~\eqref{eq:ensemble_prob}
            \STATE \textcolor[HTML]{0671b9}{ \texttt{// Sample next token}}
            \STATE $y_h \sim P(y_h|\bm{x}, \bm{y}_{<h})$
            \STATE $\bm{z}_{<t+1} \leftarrow \bm{z}_{<t} \oplus y_h$
        \ENDFOR
    \ENDFOR
    \RETURN $\bm{y}$
    \end{algorithmic}
    \end{algorithm}

\begin{algorithm}[h!]
        \renewcommand{\arraystretch}{1}
        \caption{Training Process for \method}
        \label{alg:training}
        \begin{algorithmic}
        \STATE \textbf{Input:} Base models $\{M_1, \ldots, M_K\}$, initial policy $\pi_\phi$, dataset $\mathcal{D}$
        \STATE \textbf{Output:} Trained policy $\pi_\phi$
        \FOR{each epoch}
            \STATE Initialize buffer $\mathcal{B} \leftarrow \emptyset$
            \FOR{prompt $\bm{x}$ in $\mathcal{D}$}
                \STATE Initialize $\bm{y}_0 \leftarrow \emptyset$, trajectory $\tau \leftarrow \emptyset$
                \FOR{each span $t$}
                    \STATE $s_t \leftarrow (\bm{x}, \bm{z}_{<t})$, $a_t \sim \pi_\phi(\cdot|s_t)$
                    \STATE \textcolor[HTML]{0671b9}{ \texttt{// Sample Spans}}
                    \STATE Generate span using Algorithm~\ref{alg:ensemble}
                    \STATE Compute reward $r_t$
                    \STATE Add $(s_t, a_t, r_t)$ to trajectory $\tau$
                \ENDFOR
                \STATE Add $\tau$ to buffer $\mathcal{B}$
            \ENDFOR
            \STATE Update policy using Eq.~\eqref{eq:ppo_objective} or Eq.~\eqref{eq:mappo_objective}
        \ENDFOR
        \RETURN $\pi_\phi$
        \end{algorithmic}
    \end{algorithm}
    
\newpage
\section{Hyperparameters and Computational Resources}
\label{sec:hyperparameters}

The hyperparameters used in our experiments are shown in Table~\ref{tab:hyperparameters}. All experiments are conducted on 8 NVIDIA A100 GPUs.

\begin{table}[ht]
    \centering
    \resizebox{\columnwidth}{!}{%
    \begin{tabular}{ccc}
    \toprule
    \textbf{Hyperparameter} & \textbf{Explanation} & \textbf{Values} \\ \midrule
    $\epsilon$ & Clip range & 0.2 \\
    $\gamma$ & Key GAE parameter & 0.99 \\
    $\lambda$ & Key GAE parameter & 0.95 \\
    $L$ & Span length & 4 \\
    $|B|$ & Buffer size & 128 \\
    \texttt{lr} & Maximum learning rate & 1e-4 \\
    \texttt{lr\_scheduler} & Learning rate schedule & cosine \\
    \texttt{num\_epochs} & Number of training epochs & 3 \\
    \texttt{entropy\_coef} & Entropy coefficient & 0.01 \\
    \bottomrule
    \end{tabular}
    }
    \caption{Hyperparameters used in our experiments.}
    \label{tab:hyperparameters}
    \vspace{-1em}
\end{table}

\section{Details of Value Network}
\label{sec:value_network}
The value function (critic) plays a crucial role in reinforcement learning, as it estimates the expected return of the current state. In our work, we adopt a shared network architecture~\citep{huang202237} where both the value network and policy network utilize the same DeBERTa-v3-large model. Then \method builds a value head and a policy head that share the output of the DeBERTa-v3-large model.
The value network is trained to minimize the mean squared error (MSE) loss between its predictions and the target values:
\begin{equation}
    \mathcal{L}_V =  \sum_{(s_t, a_t, r_t)} (V_\phi(s_t) - \hat{V}_t)^2
\end{equation}
where $V_\phi(s_t)$ is the value prediction for state $s_t$, and $\hat{V}_t=\sum_{l=0}^{T-t} \gamma^l r_{t+l}$ is the target value computed by the reward function $r_t$.

\section{Vocabulary Mismatch}
\label{sec:vocabulary_mismatch}
We adopt the approach proposed in \textsc{GaC}~\citep{yu-etal-2024-breaking} to address this vocabulary mismatch, which projects probability vectors from multiple LLMs into a unified vocabulary space through a mapping matrix. Specifically, we construct a comprehensive unified vocabulary set $V_u$ by aggregating tokens from each base model in our ensemble. During the construction of this unified vocabulary, we eliminate any duplicate tokens while preserving all unique tokens present across the different model vocabularies. For tokens that exist in the unified set $V_u$ but are absent from a particular model's vocabulary $V_k$, we implement a principled zero-probability assignment strategy, where the generation probability of such tokens from model $M_k$ is explicitly set to 0.
This principled approach allows us to aggregate outputs from different models at each generation step to select the next token, ensuring comprehensive coverage of the token space while maintaining proper probability distributions across models with heterogeneous tokenization schemes.
Recently, \textsc{UniTE}~\citep{yao2025determinethenensemble} proposes to match the vocabulary by considering only the union of top-k tokens from each model. This approach eliminates the need for full vocabulary alignment and reduces computational overhead, while being complementary to our method.

\section{Markov Game}
\label{sec:markov_game}
For the multi-agent setting, we extend beyond the Markov Decision Process (MDP) framework and formulate the ensemble problem as a Markov Game (MG)~\citep{littman1994markov}. A MG generalizes MDP to multiple interacting agents and is formally defined by a tuple $\langle \mathcal{K}, \mathcal{S}, \mathcal{A}, \mathcal{P}, \mathcal{R}, \gamma \rangle$, where:

\begin{itemize}
    \setlength{\itemsep}{0pt}
    \item $\mathcal{K}=\{1,\dots,K\}$ is the set of agents, with each agent corresponding to ensemble weight.
    \item $\mathcal{S}$ is the state space, representing the current prompt and generation history.
    \item $\mathcal{A}$ is the action space, which determines the ensemble weight assigned to the corresponding LLM's output.
    \item $\mathcal{P}: \mathcal{S} \times \mathcal{A} \times \mathcal{S} \rightarrow [0,1]$ is the state transition probability function.
    \item $\mathcal{R}: \mathcal{S} \times \mathcal{A} \times \mathcal{S} \rightarrow \mathbb{R}$ is the reward function for all agents.
    \item $\gamma \in [0,1]$ is the discount factor.
\end{itemize}

At each step, all agents observe the current state $s_t$ and simultaneously select actions $a_k$ according to their policies $\pi_k$. The environment then transitions to a new state $s_{t+1}$ based on the joint actions, and each agent receives its individual reward $r$. The goal of each agent is to maximize its expected discounted return:

\begin{equation}
    J_k(\pi_k) = \mathbb{E}_{\pi_k,\pi_{-k}}\left[\sum_{t=0}^{\infty} \gamma^t r_t\right]
\end{equation}
where $\pi_{-k}$ denotes the joint policy of all agents except agent $k$. This formulation enables cooperative behavior among agents through the reward design, while allowing each agent to learn specialized ensemble policies based on their corresponding LLM's strengths.

\section{Additional Experimental Results on Generalization}
\label{sec:add_results}

We provide additional experimental results on the generalization in Tables~\ref{tab:generalize1}.
The results show that the RL-assisted framework can achieve better generalization on different tasks.
\begin{table}[ht]
    \centering
    \renewcommand{\arraystretch}{1.1}
    \resizebox{\columnwidth}{!}{%
    \begin{tabular}{@{}cccc@{}}
    \toprule
    \textbf{Dataset}    & \textbf{\textsc{PairRanker}}  & \textbf{\methodppo}   & \textbf{\methodmappo}   \\ \midrule
    \textbf{GPQA}      & 32.7        & 35.1        & 34.7        \\
    \textbf{ARC-C$\,\to\,$GPQA} & 27.3 (-5.4) & \textbf{33.3 (-1.8)} & \textbf{32.6 (-2.1)} \\ \bottomrule
    \end{tabular}
    }
    \caption{Generalization from ARC-C to GPQA.}
    \label{tab:generalize1}
    \vspace{-1em}
    \end{table}

\section{Ensemble Weights Visualization}
\label{sec:visualization_ensemble_weights}

\begin{figure}[ht]
    \centering
    \includegraphics[width=\linewidth]{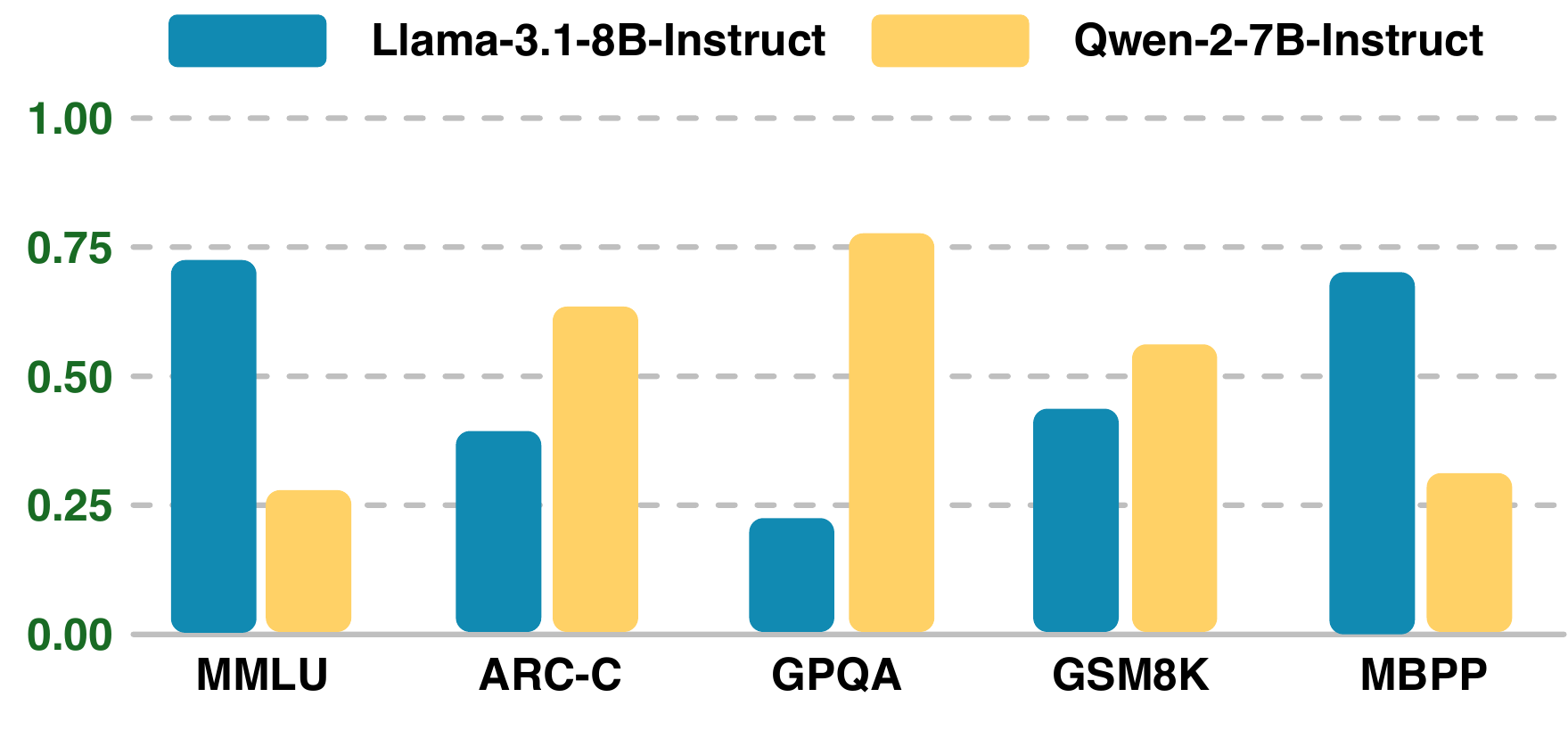}
    \caption{Ensemble weights across different benchmarks. We calculate the average weights across different benchmarks.}
    \label{fig:weight_visualization}
\end{figure}

\begin{figure*}[ht]
    \centering
    \includegraphics[width=\linewidth]{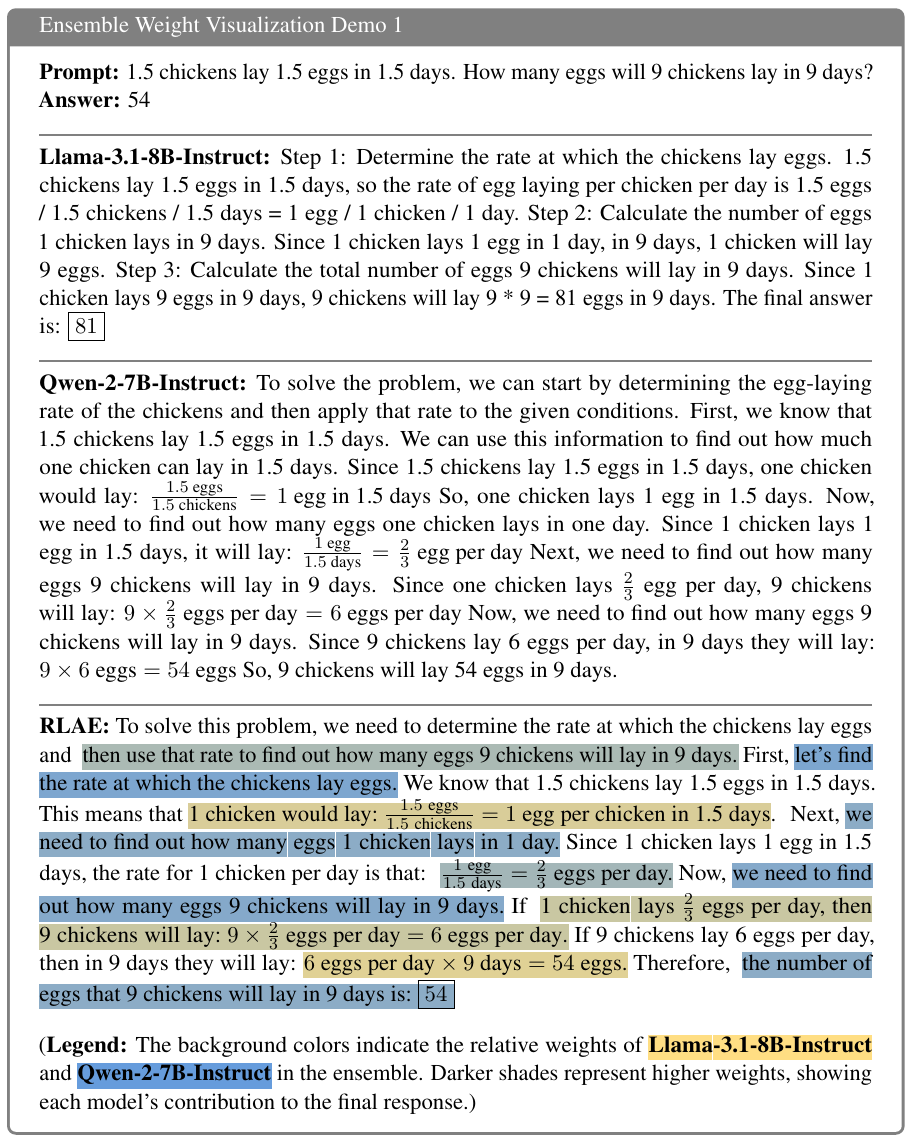}
    \caption{Visualization demo 1 of ensemble weights in a single response.}
    \label{fig:demo1}
\end{figure*}

\begin{figure*}[ht]
    \centering
    \includegraphics[width=\linewidth]{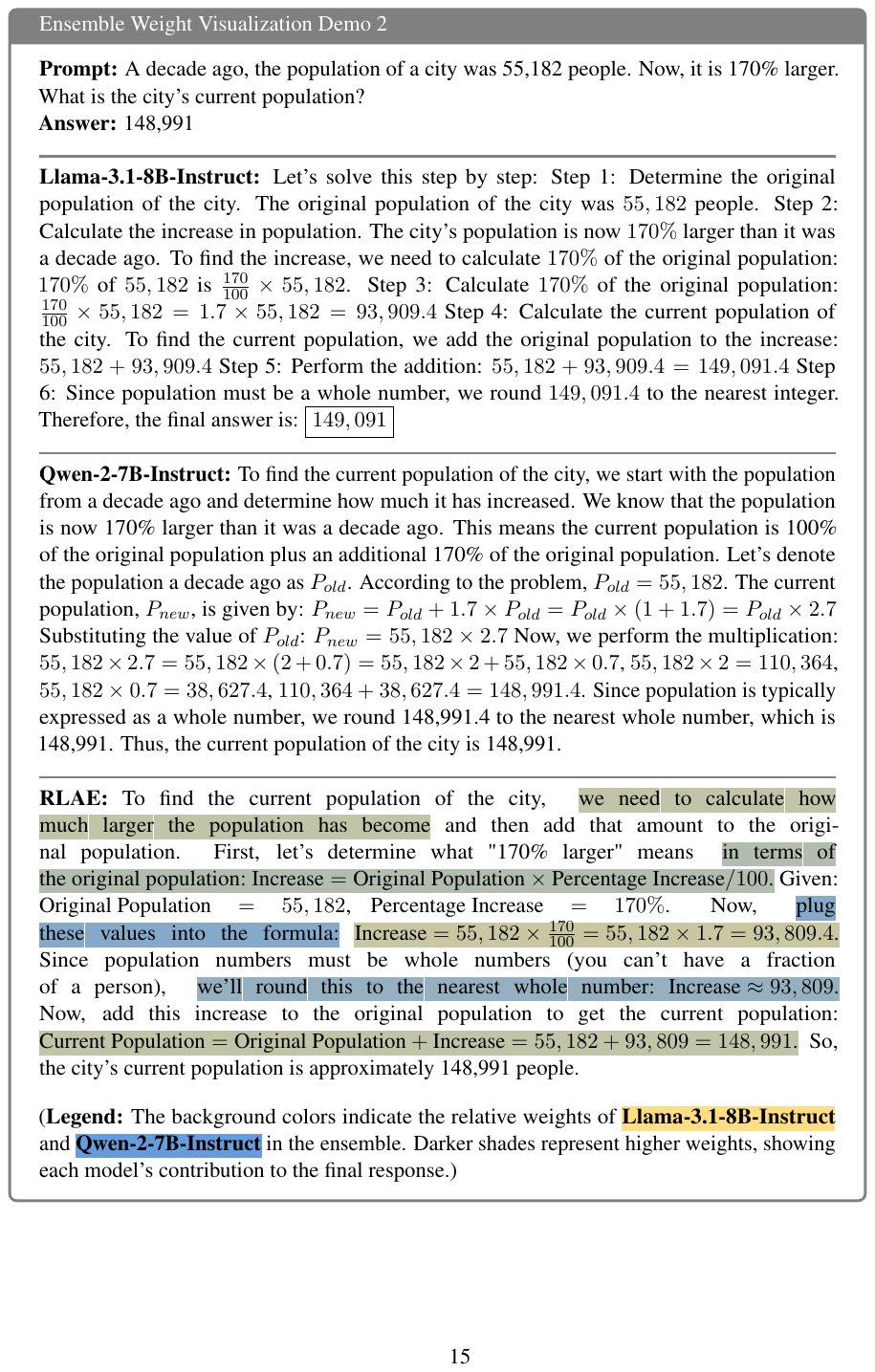}
    \caption{Visualization demo 2 of ensemble weights in a single response.}
    \label{fig:demo2}
\end{figure*}

In this section, we provide the visualizations of the ensemble weights across different benchmarks.
We first visualize the ensemble weight differences between \textit{different benchmarks}, as shown in Figure~\ref{fig:weight_visualization}.
The experimental results demonstrate that different benchmarks require distinct ensemble weights. For instance, Llama-3.1-8B-Instruct achieves higher weights on MMLU, while Qwen-2-7B-Instruct obtains higher weights on STEM-related tasks like GPQA and GSM8K.
Through reinforcement learning training, our proposed method \method can adaptively adjust ensemble weights to achieve better performance across different benchmarks.
Then, we provide a demo of the ensemble weights visualization \textit{in a single response}, as shown in Figures~\ref{fig:demo1} and \ref{fig:demo2}.
We conducted the visualization on two different prompts, and the experimental results show that \method can leverage the advantages of different models by adjusting ensemble weights, thus achieving better performance on different tasks.

\end{document}